\documentclass[conference]{IEEEtran}
\IEEEoverridecommandlockouts
\usepackage{cite}
\usepackage{amsmath,amssymb,amsfonts}
\usepackage{algorithmic}
\usepackage{multirow}
\usepackage{booktabs}
\usepackage{graphicx}
\usepackage{grffile}
\usepackage{float}
\usepackage[colorlinks, linkcolor=blue, anchorcolor=blue, citecolor=blue]{hyperref}
\hypersetup{
    colorlinks = true,
    citecolor  = blue,
    linkcolor  = blue,
    urlcolor   = blue,
}
\usepackage{textcomp}
\usepackage{xcolor}
\usepackage{comment}
\def\BibTeX{{\rm B\kern-.05em{\sc i\kern-.025em b}\kern-.08em
    T\kern-.1667em\lower.7ex\hbox{E}\kern-.125emX}}
\begin{document}

\title{OpenOpt: An Open-Source SRAM Optimizer Based on Equivalent Circuit Model}

\author{
\IEEEauthorblockN{Yikai Wang$^{1}$, Yiheng Wu$^{1}$, Can Wang$^{1}$, Bohao Liu$^{2}$, Junhao Ma$^{1}$, Zhuohua Liu $^{3}$,Qinxin Mei $^{2}$ \\ Shan Shen$^{1,*}$ }
\IEEEauthorblockA{$^{1}$\textit{Nanjing University of Science and Technology, Nanjing, 210094, China} \\
$^{2}$\textit{Shenzhen University,Shenzhen, 518060, China} \\
$^{3}$\textit{Beihang University, Beijing, 100191, China}}
\IEEEauthorblockA{Email: \{wangyikai, shanshen\}@njust.edu.cn}

\thanks{
This work is supported by the National Natural Science Foundation of China (NSFC) under grant No. 62204141 and the Fundamental Research Funds for the Central Universities under grant Nos. 30925010605 and 30924012004. $^*$ Corresponding author. }
\vspace{-20pt}
}

\maketitle

\begin{abstract}
This paper proposes a co-optimization framework that jointly optimizes SRAM architecture and transistor sizing using equivalent circuit models. The framework simplifies inactive SRAM cells into equivalent RC loads and static power models, achieving up to 61.4$\times$ simulation speedup while maintaining high fidelity (read/write delay error $<$0.22\%, power error $<$1.68\%). A joint search space encompassing architecture parameters and device sizing integrates seven algorithms including SA, PSO, Bayesian Optimization variants, and multi-objective evolutionary algorithms. Based on FreePDK45, ablation experiments confirm complementary gains from architecture selection and transistor sizing. Among all algorithms, MOEA/D achieves the best Figure of Merit (8.2721), yielding 6.2\% improvement in SNM, 73.6\% reduction in area, and 42.3\% reduction in peak power. The framework is publicly available at \href{https://github.com/W1Y1K1/OpenOpt}{OpenOpt:URL}.
\end{abstract}


\section{Introduction}
As semiconductor process nodes continue to scale down, Static Random Access Memory (SRAM), which constitutes the largest proportion of on-chip storage in modern SoCs and AI accelerators, is critical to overall chip performance, power consumption, and yield~\cite{start}. In advanced technology nodes, transistor scaling leads to significant increases in process variation, parasitic effects, and leakage current, posing severe challenges to the trade-off between PPA (Power, Performance, Area) and yield in SRAM design~\cite{introduction1}.

In traditional analog circuit design flows, SRAM design often relies on designer experience and time-consuming manual iterations. To address increasingly complex design spaces, simulation-based optimization methods have become mainstream~\cite{simulate1}--\cite{simulate3}. However, this approach faces two major bottlenecks in large-scale SRAM array design.

The first bottleneck is \emph{simulation efficiency}. Accurately capturing second-order effects in advanced processes requires high-precision SPICE simulation. However, SRAM arrays contain millions of repetitive cells, and as the array size increases, the runtime of full-array SPICE simulation grows super-linearly. Although surrogate models have been employed to accelerate simulation~\cite{beida}, such black-box models lack physical interpretability and are prone to prediction errors under sparse sampling and boundary conditions, making it difficult to meet high-reliability design requirements. Prior analytical RC models~\cite{model} preserve physical topology but derived parasitic parameters solely from individual transistor geometry and only captured timing behavior, neglecting power modeling and inter-transistor coupling effects.

The second bottleneck is the \emph{joint optimization of architecture and bit-cell sizing}. SRAM performance depends not only on transistor sizing but also on architecture-level configuration (e.g., row/column counts and column-mux ratio)~\cite{array}. Traditional optimization methods typically decouple these two aspects or perform local searches in low-dimensional spaces, failing to find globally optimal solutions in the joint architecture-sizing space. Recent two-level frameworks such as OpenACMv2~\cite{openacmv2} decompose architecture exploration and transistor sizing into sequential stages for tractability, but such decoupling inherently limits the search to stage-wise optima rather than the joint global optimum. Existing automation tools such as OpenRAM~\cite{openram} support architecture generation but lack integrated support for transistor sizing or optimization algorithms. Conversely, OpenYield~\cite{openyield} focuses on transistor sizing without addressing architecture-level exploration.

To address these challenges, this work proposes a large-scale SRAM co-optimization framework based on equivalent circuit models. Built upon the high-fidelity models of OpenYield~\cite{openyield}, the main contributions include:

\begin{enumerate}
    \item We propose an equivalent circuit model that replaces inactive SRAM cells with compact RC loads, achieving up to 61.4$\times$ speedup with $<$1.68\% error.
    \item We construct a unified platform co-optimizing architecture and transistor sizing, with ablation experiments confirming that the two stages provide complementary and significant gains.
    \item We integrate five single-objective and two multi-objective optimization algorithms~\cite{sa1983}--\cite{nsga2009} and release them in open-source form, providing complete toolchain support for SRAM design optimization.
\end{enumerate}
Through this platform, we systematically explore the joint architecture-transistor design space of SRAM, achieving comprehensive optimization of stability, power consumption, area, and delay.

\section{Background}
\subsection{SRAM Simulation and Optimization}

Simulation acceleration is critical for large-scale SRAM design. Neural network surrogate models~\cite{beida} achieve speedup by fitting nonlinear mappings but lack physical interpretability and suffer accuracy degradation under sparse sampling and boundary conditions. Analytical RC equivalent models~\cite{model} preserve physical topology and require no training data, but prior work derived parasitic parameters from individual transistor geometry, addressed only timing metrics without modeling power, and neglected leakage current of inactive cells. Since the vast majority of cells remain unselected during read/write operations and affect performance only through parasitic loading, simplifying these cells with equivalent models can achieve significant acceleration while maintaining accuracy even under extreme process corners.

SRAM optimization has been studied from both architecture and transistor perspectives. Architecture-level tools such as OpenRAM~\cite{openram} support automated layout generation with configurable row/column counts and mux ratios, but do not integrate transistor sizing or optimization algorithms. Conversely, OpenYield~\cite{openyield} focuses on variation-aware transistor sizing for bitcell stability and yield without architecture-level exploration. OpenACMv2~\cite{openacmv2} adopts a two-level strategy that decouples architecture search from device sizing for tractability, but this sequential decomposition cannot guarantee joint global optimality. More broadly, simulation-based optimization approaches~\cite{simulate1}--\cite{simulate3} have become mainstream for navigating complex design spaces, yet scaling them to joint architecture-sizing search with large SRAM arrays remains an open challenge.

\subsection{Problem Formulation}
The exploration of the SRAM design space can be formulated as a constrained multi-objective black-box optimization problem. The primary objective is to identify an optimal set of design variables $\mathbf{x}$ such that PPA (Power, Performance, Area) and stability metrics achieve Pareto optimality while satisfying process design rules and functional constraints. Mathematically, this problem is expressed as:
\begin{equation}
\begin{aligned}
\min_{\mathbf{x} \in \Omega} \quad & \mathbf{F}(\mathbf{x}) = \left[ f_1(\mathbf{x}), f_2(\mathbf{x}), \dots, f_m(\mathbf{x}) \right]^T \\
\text{s.t.} \quad & f_j(\mathbf{x}) \leq y_j, \quad j=1,\dots,p
\end{aligned}
\end{equation}
where $\mathbf{x}$ represents a mixed design variable vector composed of discrete architecture parameters and continuous device sizing parameters; $\mathbf{F}(\mathbf{x})$ is the objective function vector mapping key circuit metrics such as power, performance, area, and stability; and $\mathbf{y}$ represents the circuit design constraints. To address the challenges of multi-objective trade-offs in SRAM circuit design, this paper integrates various types of optimization algorithms to accommodate different search requirements.

\begin{figure}[tb]
\centering
\includegraphics[width=\linewidth]{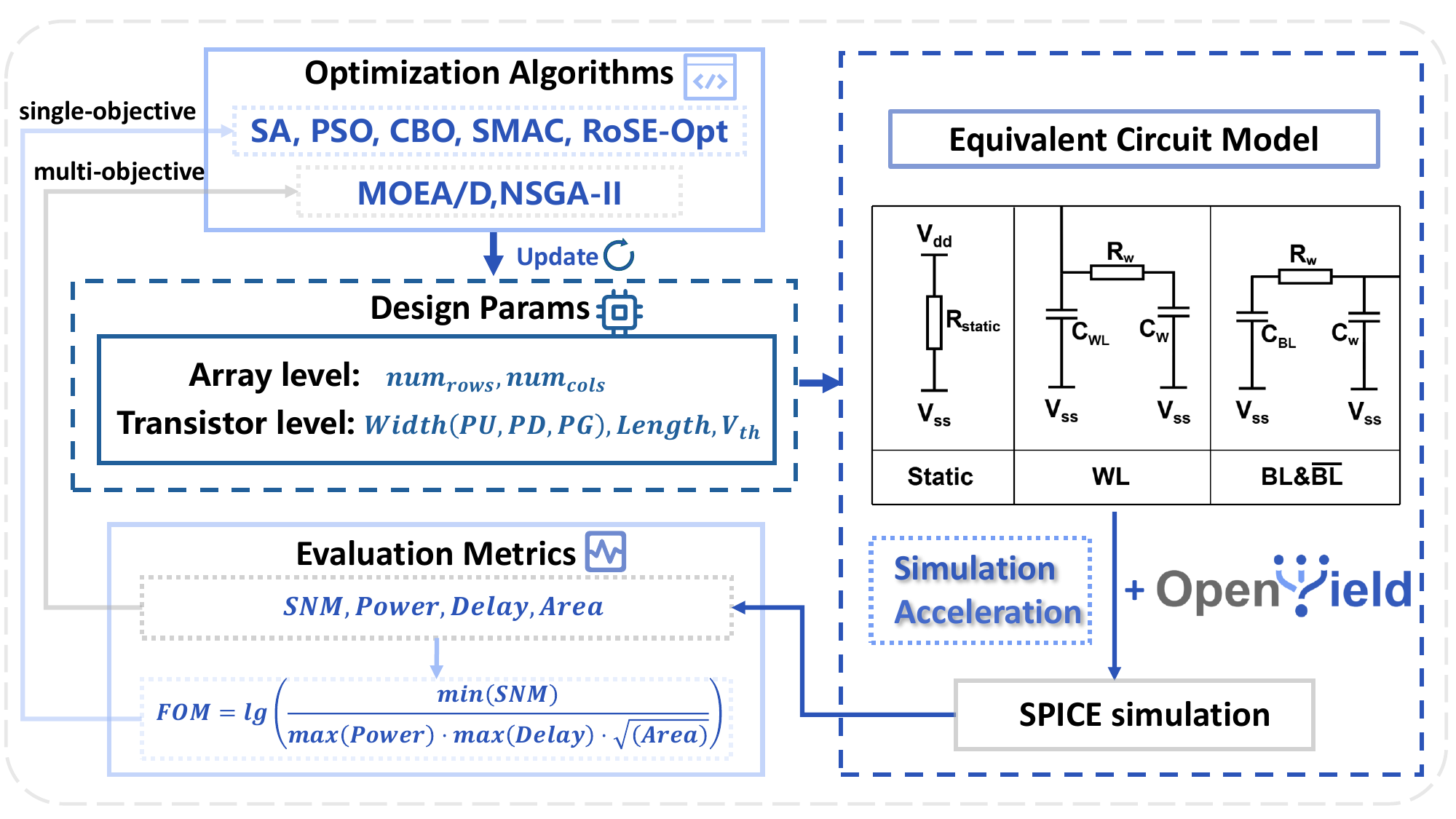}
\caption{OpenOpt framework overview, comprising three core modules: equivalent circuit model for simulation acceleration, optimization algorithms (single-objective SA/PSO/CBO/SMAC/RoSE-Opt and multi-objective NSGA-II/MOEA/D), and circuit generator with SPICE simulator based on OpenYield.}
\label{fig:flowchart}
\end{figure}

\section{Proposed Method}
The experimental workflow is illustrated in Fig.~\ref{fig:flowchart}. OpenOpt first designs specific test circuits, extracts key parasitic parameters from SRAM cell-level simulations, and constructs the equivalent circuit model. Subsequently, optimization algorithms generate candidate designs within the joint architecture-sizing search space, and the system evaluates each candidate via SPICE simulation. For single-objective algorithms, a comprehensive Figure of Merit (FoM) is calculated and fed back to refine the search strategy; for multi-objective algorithms (NSGA-II, MOEA/D), the raw performance metrics (SNM, Power, Delay, Area) are directly provided to guide Pareto front exploration. This iterative process continues until the termination criterion is reached.

\subsection{SRAM Cell Equivalent Circuit Design}

\begin{figure}[t]
\centering
\includegraphics[width=1.00\linewidth]{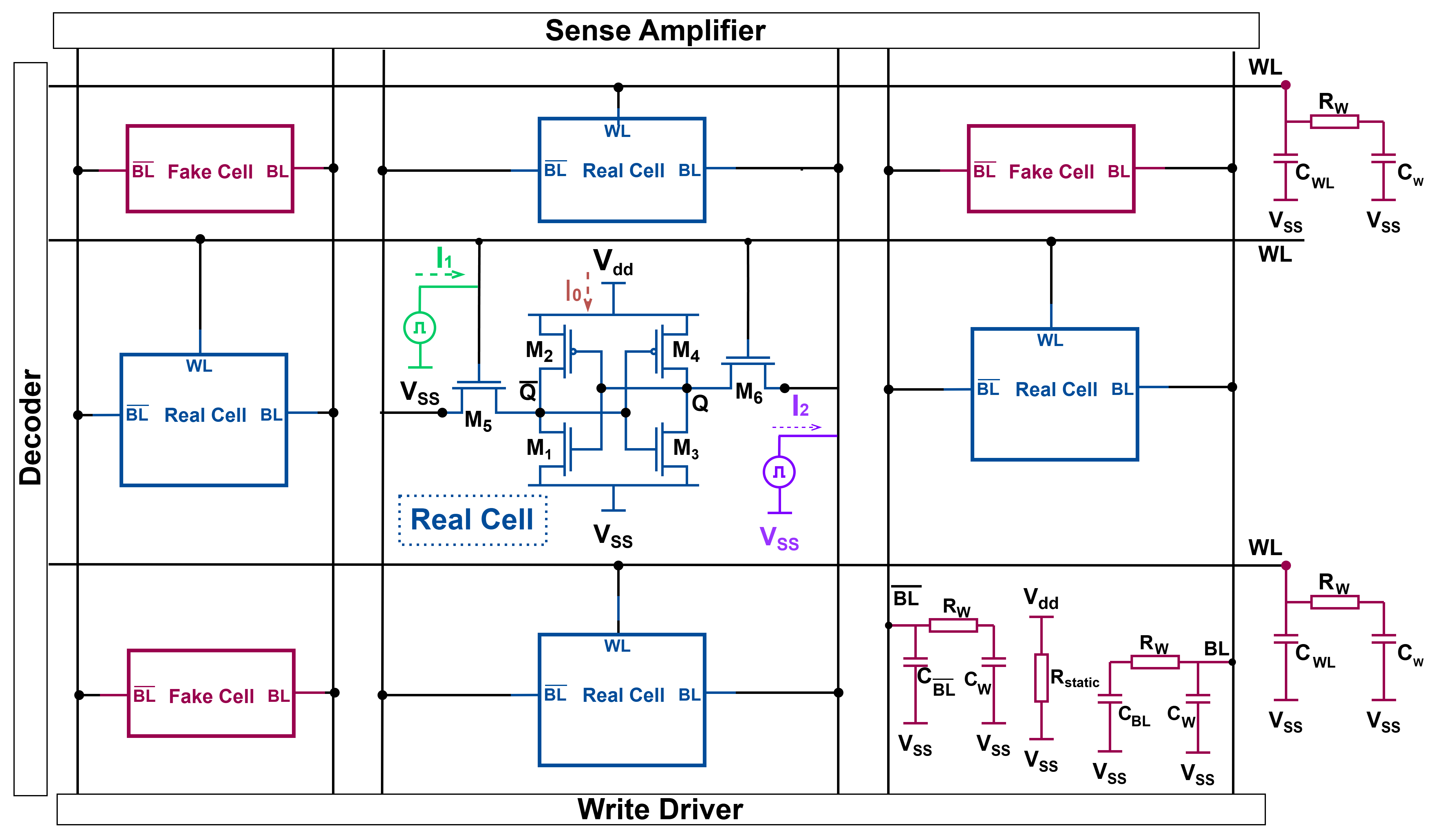}
\caption{Schematic diagram of 3$\times$3 array with equivalent circuits. The center cell is a complete 6T SRAM for testing, while surrounding cells are replaced by equivalent RC models.}
\label{fig:3x3}
\end{figure}

We adopt a hierarchical approach: first building a compact 3$\times$3 array model with a complete target cell surrounded by equivalent circuit elements, then extending it to arbitrary array sizes through linear superposition. As shown in Fig.~\ref{fig:3x3}, a complete 6-transistor SRAM cell is placed at the center as the target cell, while the surrounding 8 positions are replaced by equivalent circuit models. The array is equipped with complete peripheral circuits: the row decoder drives the selected WL to high level; the write driver transmits data through bit line differential pairs; the sense amplifier detects and amplifies weak voltage differences during read operations; and the precharge circuit precharges bit lines before each access.

Each equivalent cell model consists of four components: BL terminal capacitance, BLB terminal capacitance, WL terminal capacitance, and static power resistance.

\subsubsection{Equivalent Model for BL and BLB Terminals}
\begin{figure}[t]
\centering
\begin{minipage}[t]{0.5\linewidth}
\centering
\includegraphics[width=\linewidth]{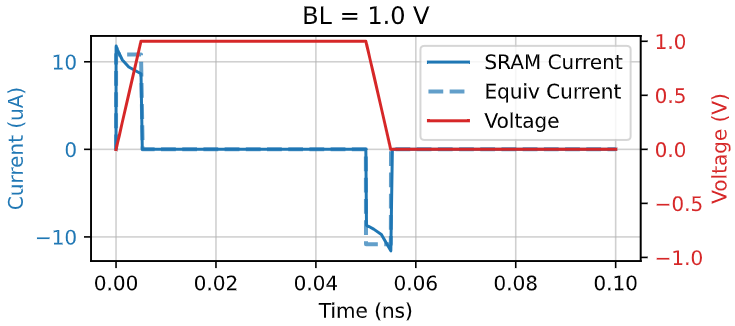}\\
\footnotesize (a) BL terminal
\end{minipage}\hfill
\begin{minipage}[t]{0.5\linewidth}
\centering
\includegraphics[width=\linewidth]{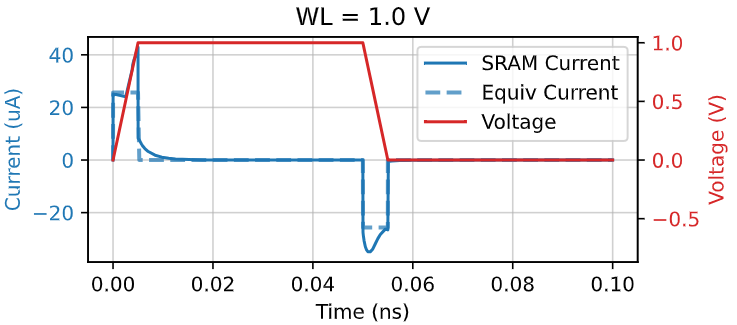}\\
\footnotesize (b) WL terminal
\end{minipage}
\caption{Equivalent-model fit for (a) BL and (b) WL terminals. Model (green) matches the measured current response (blue).}
\label{fig:bl_wl_test}
\end{figure}

For cells in unselected rows, WL is grounded and the access transistor is off, so the BL/BLB input current is proportional to $dV/dt$, exhibiting pure capacitive behavior. As shown in Fig.~\ref{fig:bl_wl_test}(a), the equivalent capacitance $C_{\mathrm{BL}}$ (or $C_{\mathrm{BLB}}$) achieves excellent agreement with actual simulation results. When the access transistor is completely turned off, the internal storage state has negligible impact on external behavior, allowing us to ignore internal state differences of inactive cells and significantly reduce modeling complexity.

\subsubsection{Equivalent Model for WL Terminal}
The WL terminal exhibits more complex piecewise behavior: pure capacitive at lower voltages, with an additional RC network above the access transistor threshold (Fig.~\ref{fig:bl_wl_test}(b)). We adopt a simplified single-capacitance model $C_{\mathrm{WL,simple}}$ that provides adequate accuracy with superior computational efficiency.

\subsubsection{Static Power Equivalent Model}
After circuit stabilization, the $V_{\mathrm{DD}}$ terminal current remains essentially constant, modeled as $R_{\mathrm{static}} = V_{\mathrm{DD}}/I_{\mathrm{static}}$ obtained from cell-level simulation.

\subsubsection{Array Extension}
The 3$\times$3 model extends to arbitrary $m \times n$ arrays through linear superposition:
\begin{equation}
C_{\mathrm{BL,total}} = (m-1) \times C_{\mathrm{BL,unit}}
\end{equation}
\begin{equation}
C_{\mathrm{WL,total}} = (n-1) \times C_{\mathrm{WL,unit}}
\end{equation}
\begin{equation}
R_{\mathrm{static,total}} = \frac{R_{\mathrm{static,unit}}}{m \times n - 1}
\end{equation}
This ensures computational complexity is decoupled from array size, achieving significant simulation acceleration.

\subsection{Optimization Algorithm Design Space}
After incorporating the equivalent circuit model for simulation acceleration, the focus shifts to SRAM macro design optimization through a two-level co-optimization strategy: bank-level configuration under capacity constraints, and transistor-level sizing.

\subsubsection{Design Parameters}
We conduct searches on both architecture parameters and sizing parameters under capacity constraints, with the goal of determining an array organization and optimal transistor sizing---described by the number of rows $r$, number of columns $c$, multiplexing ratio $\mu$, number of arrays $n_a$, transistor dimensions $W/L$, and transistor type $\gamma$---to balance dynamic performance with memory bank area. All feasible configurations satisfy the capacity constraint:
\begin{equation} 
r \times c \times n_a = \text{Capacity}
\end{equation}
where $\text{Capacity}$ represents the total number of storage bits. This paper sets $\text{Capacity}=32\,\mathrm{KB}\times 8=262{,}144$ bits, and determines $n_a$ accordingly to ensure full capacity utilization. Candidate values use a sparse grid with step size of 2: $r\in\{8,16,\ldots,512\}$, $c\in\{8,16,\ldots,256\}$, and $\mu$ is selected from a set given by process and design specifications.

\subsubsection{Architecture-Level Penalty Model}
\label{sec:penalty}
When the optimizer partitions the total capacity into $n_a$ sub-arrays, additional peripheral logic is required for chip-select decoding and column multiplexing. We model the resulting delay and power penalties analytically, based on two physically grounded observations.

Chip-select delay. To address $n_a$ sub-arrays, a binary selection tree with $\lceil\log_2 n_a\rceil$ stages is required. Each stage contributes a gate delay $\tau_{\text{cs}}$, which is extracted once via SPICE simulation of a single chip-select logic gate in the standard cell library. The total chip-select overhead is therefore:
\begin{equation}
\Delta D_{\text{cs}} = \tau_{\text{cs}} \cdot \lceil\log_2 n_a\rceil
\label{eq:cs_delay}
\end{equation}
This logarithmic scaling reflects the standard depth of a binary decoder tree and is well-established in digital design~\cite{openram}.

Column multiplexer penalty. When column multiplexing is enabled ($\mu > 1$), a $\mu$-to-1 pass-gate multiplexer is inserted on each output column. Its propagation delay $\tau_{\text{mux}}$ and switching power $P_{\text{mux}}$ are similarly extracted from SPICE characterization. The read path incurs both an additional delay and power due to the multiplexer:
\begin{equation}
\Delta D_{\text{mux}} = \tau_{\text{mux}} \cdot \lceil\log_2 \mu\rceil
\label{eq:mux_delay}
\end{equation}
\begin{equation}
\Delta P_{\text{mux}} = (n_a - 1) \cdot P_{\text{mux}}
\label{eq:mux_power}
\end{equation}
where Eq.~\eqref{eq:mux_power} accounts for the leakage of unselected multiplexers across all $n_a - 1$ inactive arrays.

Multi-round access penalty. Let $c_{\text{out}}$ denote the required number of output columns (default 16). When the effective output width $c / \mu < c_{\text{out}}$, a single access cannot fulfill the full word width, and $k = \lfloor c_{\text{out}} \cdot \mu / c \rfloor$ sequential access rounds are needed. The total delay scales linearly:
\begin{equation}
D_{\text{total}} = D_{\text{base}} \times k
\label{eq:multi_round}
\end{equation}
where $D_{\text{base}}$ includes the array read/write delay plus $\Delta D_{\text{cs}}$ and $\Delta D_{\text{mux}}$. The dynamic power accumulates proportionally as each round activates the sense amplifiers and write drivers independently.

Composite penalty. Combining the above terms, the corrected read delay is:
\begin{equation}
D'_{rd} = \left(D_{rd} + \Delta D_{\text{cs}} + \Delta D_{\text{mux}}\right) \times k
\end{equation}
and similarly for the write delay (without $\Delta D_{\text{mux}}$). The corrected power metrics are:
\begin{equation}
P'_{rd} = P_{rd} + \Delta P_{\text{mux}} + P_{r,\text{dyn}} \times k
\end{equation}
\begin{equation}
P'_{wr} = P_{wr} + P_{w,\text{dyn}} \times k
\end{equation}

This penalty model effectively penalizes configurations that blindly increase $n_a$ (incurring logarithmic delay overhead) or reduce array column count below $c_{\text{out}}$ (incurring multi-round access cost), thereby guiding the optimizer toward a balanced architecture.

\subsubsection{Objective Function}
For single-objective optimization algorithms (SA, PSO, CBO, SMAC, RoSE-Opt), we define a comprehensive Figure of Merit that balances all performance metrics:
\begin{equation}
{\small \text{FoM}=\log_{10}\!\left(\frac{\text{SNM}_{\min}}{P_{\max}\times D_{\max}\times \sqrt{\text{Area}}}\right)}
\label{FOM}
\end{equation}
where $D_{\max}=\max(D'_{rd},D'_{wr})$ and $P_{\max}=\max(P'_{rd},P'_{wr})$. A larger FoM indicates better overall efficiency, emphasizing low power, small area, short access delay, and large noise margin.

For multi-objective optimization algorithms (NSGA-II, MOEA/D), the raw performance metrics---SNM, Power, Delay, and Area---are directly provided as optimization objectives, allowing these algorithms to explore the Pareto front without scalar aggregation.

\subsection{Optimization Algorithms}
OpenOpt integrates seven algorithms in a unified loop: at each iteration, the optimizer proposes a candidate design, evaluates it via the equivalent-circuit-accelerated SPICE simulation, and updates its search strategy.

\subsubsection{Single-objective algorithms}
These algorithms maximize the scalar FoM (Eq.~\eqref{FOM}):
(i)~\textit{SA}~\cite{sa1983} perturbs the current design by $\mathbf{x}'=\mathbf{x}_t+\mathcal{N}(0,\sigma(T_t))$ and accepts worse solutions with Metropolis probability $\exp(-\Delta f/T_t)$; the temperature follows $T_{t+1}=\alpha T_t$ ($\alpha{=}0.98$) with a restart after 50 stagnant evaluations.
(ii)~\textit{PSO}~\cite{pso1995} updates each particle's velocity via $\mathbf{v}_i^{t+1}=w\mathbf{v}_i^t+c_1\mathbf{r}_1\!\odot\!(\mathbf{p}_i-\mathbf{x}_i^t)+c_2\mathbf{r}_2\!\odot\!(\mathbf{g}-\mathbf{x}_i^t)$, combining personal best $\mathbf{p}_i$ and global best $\mathbf{g}$ with Gaussian jitter and stochastic reinitialization to avoid premature convergence.
(iii)~\textit{CBO}~\cite{cbo2014} fits a Gaussian Process surrogate and selects candidates by maximizing constrained Expected Improvement $\text{EI}(\mathbf{x})\!\cdot\!\Pr(\text{feasible}\mid\mathbf{x})$.
(iv)~\textit{SMAC}~\cite{smac2011} uses a Random Forest surrogate instead, natively handling the mixed continuous-categorical design space.
(v)~\textit{RoSE-Opt}~\cite{RoSE_Opt2} couples GP-based Bayesian optimization with a PPO reinforcement learning agent that learns an adaptive sampling policy from accumulated evaluations.

\subsubsection{Multi-objective algorithms}
MO algorithms directly optimize the four raw metrics---SNM, Power, Delay, and Area---to approximate the Pareto front $\mathcal{F}^*$:
(i)~\textit{MOEA/D}~\cite{moead2007} decomposes the problem into $N$ scalar subproblems via weight vectors $\{\boldsymbol{\lambda}_k\}$, each minimizing $g^{\text{te}}(\mathbf{x}\mid\boldsymbol{\lambda},\mathbf{z}^*)=\max_i\{\lambda_i|f_i(\mathbf{x})-z_i^*|\}$, with neighboring subproblems sharing offspring to promote Pareto diversity.
(ii)~\textit{NSGA-II}~\cite{nsga2009} ranks the combined parent-offspring population by fast non-dominated sorting and crowding distance, preserving well-spread solutions across the trade-off landscape.

\section{Experimental Results and Analysis}

\subsection{Equivalent Circuit Experiments}
\begin{figure}[H]
\centering
\begin{minipage}[t]{0.70\linewidth}
\centering
\includegraphics[width=\linewidth]{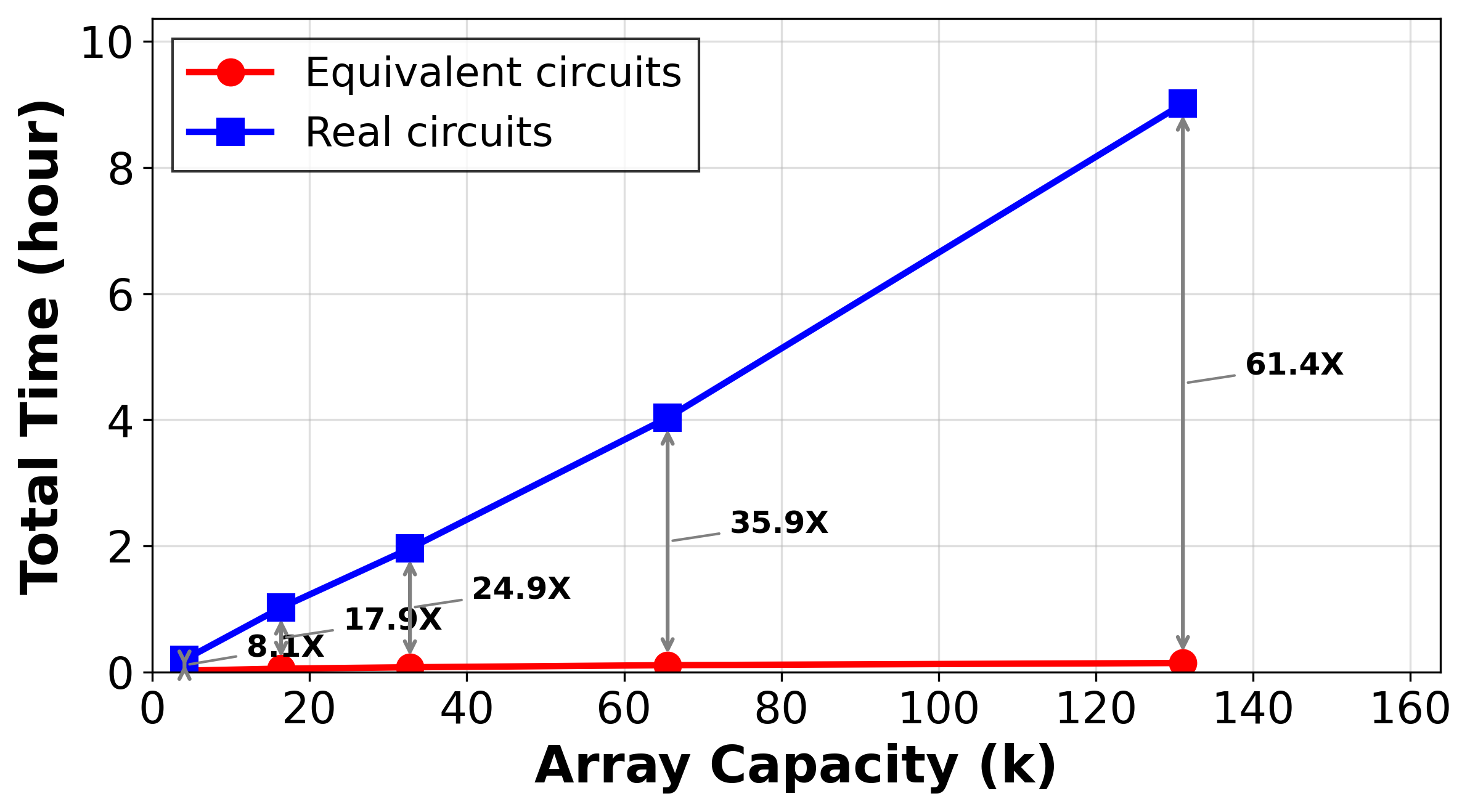}
\end{minipage}
\hfill
\caption{Simulation speedup ratios across array combinations from 32$\times$16 to 256$\times$512.}
\label{fig:speedup_ratio}
\end{figure}

The equivalent circuit model demonstrates significant simulation acceleration across SRAM blocks of different scales, as shown in Fig.~\ref{fig:speedup_ratio}. The speedup is particularly pronounced for larger blocks: medium-scale arrays achieve up to 30$\times$ speedup, and the large 256$\times$512 configuration reaches 61.4$\times$. This trend arises because the number of actual SRAM cells grows quadratically with row and column counts, whereas the equivalent circuit model replaces all cells outside the target row and column, maintaining essentially linear growth in the number of simulated components.

\begin{table}[H]
\caption{Simulation accuracy comparison between full SRAM circuits (Real) and equivalent circuits (Equiv.) of different array configurations.}
\centering
\scriptsize
\setlength{\tabcolsep}{3pt}
\renewcommand{\arraystretch}{1.1}
\resizebox{\columnwidth}{!}{%
\begin{tabular}{l |ccc| ccc| ccc} 
\hline
\multirow{2}{*}{\textbf{Metric}} & \multicolumn{3}{c|}{\textbf{32$\times$16}} & \multicolumn{3}{c|}{\textbf{128$\times$64}} & \multicolumn{3}{c}{\textbf{512$\times$512}} \\ 
\cline{2-10}
 & Equiv. & Real & Error\% & Equiv. & Real & Error\% & Equiv. & Real & Error\% \\
\hline
$D_{rd}$     & 5.71e-10 & 5.71e-10 & -0.05\% & 1.47e-09 & 1.47e-09 & -0.03\% & 4.99e-09 & 4.99e-09 & -0.01\% \\
$P_{rd,avg}$ & -1.16e-04 & -1.15e-04 & 0.50\% & -1.45e-03 & -1.47e-03 & -0.86\% & -4.35e-02 & -4.39e-02 & -0.92\% \\
$P_{rd,dyn}$ & -3.17e-04 & -3.15e-04 & 0.53\% & -3.91e-03 & -3.95e-03 & -0.92\% & -1.19e-01 & -1.20e-01 & -0.97\% \\
$P_{rd,stc}$ & -2.45e-06 & -2.47e-06 & -0.83\% & -5.57e-05 & -5.54e-05 & 0.60\% & -1.10e-03 & -1.09e-03 & 0.79\% \\
$D_{wr}$     & 3.58e-10 & 3.58e-10 & 0.03\% & 4.90e-10 & 4.90e-10 & 0.06\% & 1.16e-09 & 1.16e-09 & 0.22\% \\
$P_{wr,avg}$ & -7.29e-05 & -7.38e-05 & -1.27\% & -8.66e-04 & -8.72e-04 & -0.67\% & -2.58e-02 & -2.60e-02 & -0.59\% \\
$P_{wr,dyn}$ & -1.98e-04 & -2.00e-04 & -1.28\% & -2.32e-03 & -2.34e-03 & -0.71\% & -6.92e-02 & -6.97e-02 & -0.69\% \\
$P_{wr,stc}$ & -2.25e-06 & -2.27e-06 & -0.79\% & -3.72e-05 & -3.71e-05 & 0.32\% & -1.11e-03 & -1.09e-03 & 1.68\% \\
\hline
\end{tabular}%
}
\label{tabe:precision}
\end{table}

As shown in Table~\ref{tabe:precision}, the equivalent circuit model maintains high fidelity across all array scales, with read/write delay errors within 0.22\% and power errors within 1.68\%. These results confirm that the model adequately preserves key performance metrics while providing a fast and accurate evaluation engine for iterative optimization.

\subsection{Optimization Algorithm Experiments}

\begin{table*}[!t]
\caption{Comparison of design parameters and performance metrics explored by different optimization algorithms.}
\centering
\scriptsize
\setlength{\tabcolsep}{1pt}
\renewcommand{\arraystretch}{1.1}
\begin{tabular*}{0.95\textwidth}{@{\extracolsep{\fill}}l|cccccccc}
\hline
Measure & w/o Opt. & MOEA/D & NSGA-II & SMAC & Rose\_opt & SA & CBO & PSO \\
\hline
Array Size        & $16 \times 16$    & $64 \times 32$  & $64 \times 32$  & $64 \times 32$  & $64 \times 32$  & $64 \times 32$  & $32 \times 32$  & $32 \times 32$  \\
Array Num.        & 1024              & 128             & 128             & 128             & 128             & 128             & 256             & 256             \\
Column Mux        & Off               & On              & On              & On              & On              & On              & On              & On             \\
NMOS Model        & VTG               & VTG             & VTG             & VTH             & VTG             & VTL             & VTL             & VTL             \\
PMOS Model        & VTG               & VTH             & VTH             & VTH             & VTH             & VTH             & VTH             & VTL             \\
$W_{\mathrm{PD}}$ ($\mu$m) & 0.205 & 0.072 & 0.075 & 0.108 & 0.079 & 0.072 & 0.045 & 0.045 \\
$W_{\mathrm{PU}}$ ($\mu$m) & 0.090 & 0.072 & 0.080 & 0.072 & 0.082 & 0.108 & 0.048 & 0.049 \\
$W_{\mathrm{PG}}$ ($\mu$m) & 0.135 & 0.162 & 0.158 & 0.108 & 0.154 & 0.162 & 0.107 & 0.156 \\
$L$ (nm)          & 50.00             & 50.47           & 52.08           & 40.00           & 52.03           & 60.00           & 71.85           & 70.01           \\
\hline
$T_{\mathrm{read}}$ (ns)   & 2.564  & 2.274   & 2.271   & 2.394   & 2.259   & 2.259   & 2.448   & 2.430   \\
$T_{\mathrm{write}}$ (ns)  & 0.750  & 0.638   & 0.637   & 0.626   & 0.636   & 0.628   & 0.721   & 0.679   \\
$P_{\mathrm{max}}$ (mW) & 3.331 & 1.923 & 1.889 & 1.787 & 1.887 & 2.259 & 1.658 & 1.847 \\
min SNM (V)       & 0.273  & 0.290  & 0.276  & 0.270  & 0.268  & 0.314  & 0.315  & 0.365  \\
Area ($\text{mm}^2$) & 0.4773 & 0.1253 & 0.1262 & 0.1211 & 0.1259 & 0.1338 & 0.2042 & 0.2270 \\
\hline
FoM      & 7.6653 & 8.2721 & 8.2577 & 8.2591 & 8.2482 & 8.2263 & 8.2356 & 8.2316 \\
\hline
\end{tabular*}
\label{table:opt_results}
\end{table*}

\begin{table}[!t]
\caption{Ablation: architecture vs.\ transistor sizing (MOEA/D).}
\centering
\scriptsize
\setlength{\tabcolsep}{2pt}
\renewcommand{\arraystretch}{1.05}
\resizebox{\columnwidth}{!}{%
\begin{tabular}{l|ccc|cc}
\hline
Measure & S0: Orig. & S1: Arch. & S2: Arch.+Siz. & $\Delta$(S0$\to$S1) & $\Delta$(S1$\to$S2) \\
\hline
Array Size          & $16 \times 16$   & $64 \times 32$   & $64 \times 32$   & --- & --- \\
Array Num.          & 1024             & 128              & 128              & --- & --- \\
Column Mux          & Off              & On               & On               & --- & --- \\
NMOS / PMOS         & VTG / VTG        & VTG / VTG        & VTG / VTH        & --- & --- \\
$W_{\mathrm{PD}}$ ($\mu$m) & 0.205     & 0.205            & 0.072            & --- & --- \\
$W_{\mathrm{PU}}$ ($\mu$m) & 0.090     & 0.090            & 0.072            & --- & --- \\
$W_{\mathrm{PG}}$ ($\mu$m) & 0.135     & 0.135            & 0.162            & --- & --- \\
$L$ (nm)            & 50.00            & 50.00            & 50.47            & --- & --- \\
\hline
$T_{\mathrm{read}}$ (ns)   & 2.564  & 2.319  & 2.274  & $\downarrow$9.6\%  & $\downarrow$1.9\% \\
$T_{\mathrm{write}}$ (ns)  & 0.750  & 0.670  & 0.638  & $\downarrow$10.7\% & $\downarrow$4.8\% \\
$P_{\mathrm{max}}$ (mW)    & 3.331  & 2.040  & 1.923  & $\downarrow$38.8\% & $\downarrow$5.7\% \\
min SNM (V)                & 0.273  & 0.273  & 0.290  & ---   & $\uparrow$6.3\% \\
Area ($\text{mm}^2$)       & 0.4773 & 0.1503 & 0.1253 & $\downarrow$68.5\% & $\downarrow$16.6\% \\
\hline
FoM               & 7.6653 & 8.1725 & 8.2721 & $\uparrow$6.6\%    & $\uparrow$1.2\% \\
\hline
\end{tabular}%
}
\label{table:ablation}
\begin{flushleft}
\scriptsize \textit{S0:} unoptimized baseline~\cite{openram}. \textit{S1:} MOEA/D architecture with original sizing. \textit{S2:} full co-optimization.
\end{flushleft}
\end{table}

\begin{figure*}[!t]
    \centering
    \begin{minipage}[b]{0.24\linewidth}
        \centering
        \includegraphics[width=\linewidth]{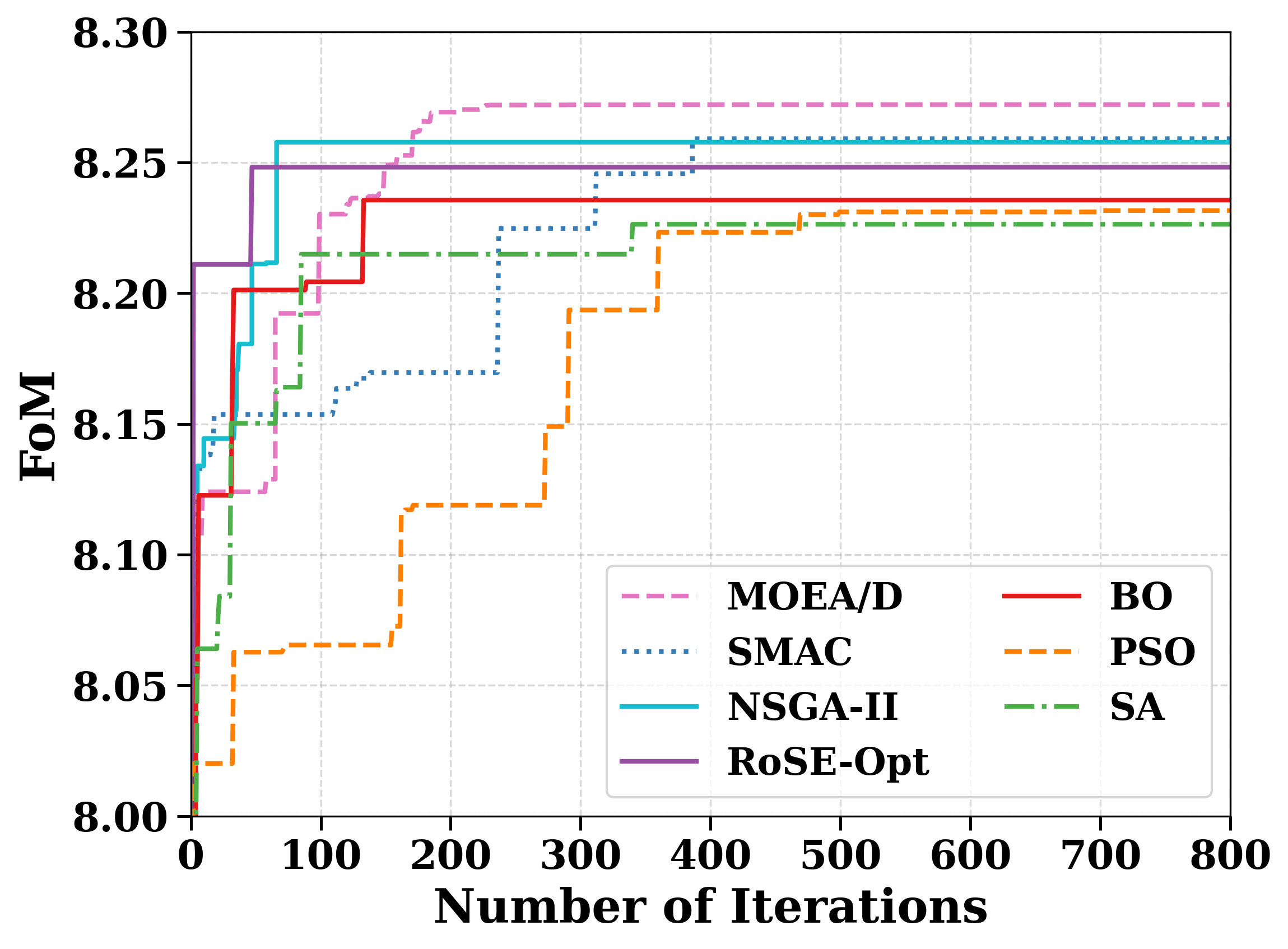} \\
        \footnotesize (a) FoM convergence curve
    \end{minipage}
    \hfill
    \begin{minipage}[b]{0.24\linewidth}
        \centering
        \includegraphics[width=\linewidth]{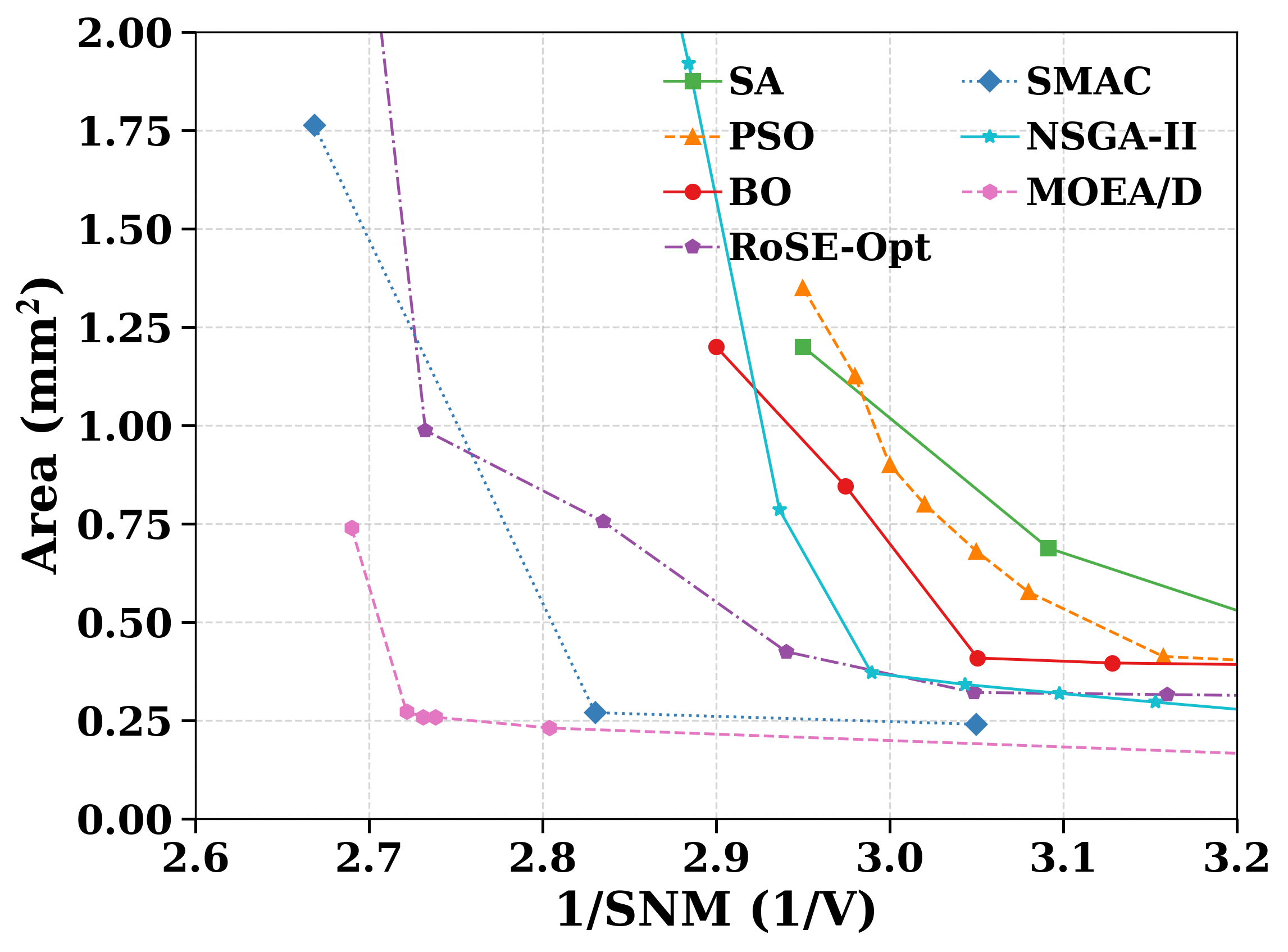} \\ 
        \footnotesize (b) Area-SNM comparison
    \end{minipage}
    \hfill
    \begin{minipage}[b]{0.24\linewidth}
        \centering
        \includegraphics[width=\linewidth]{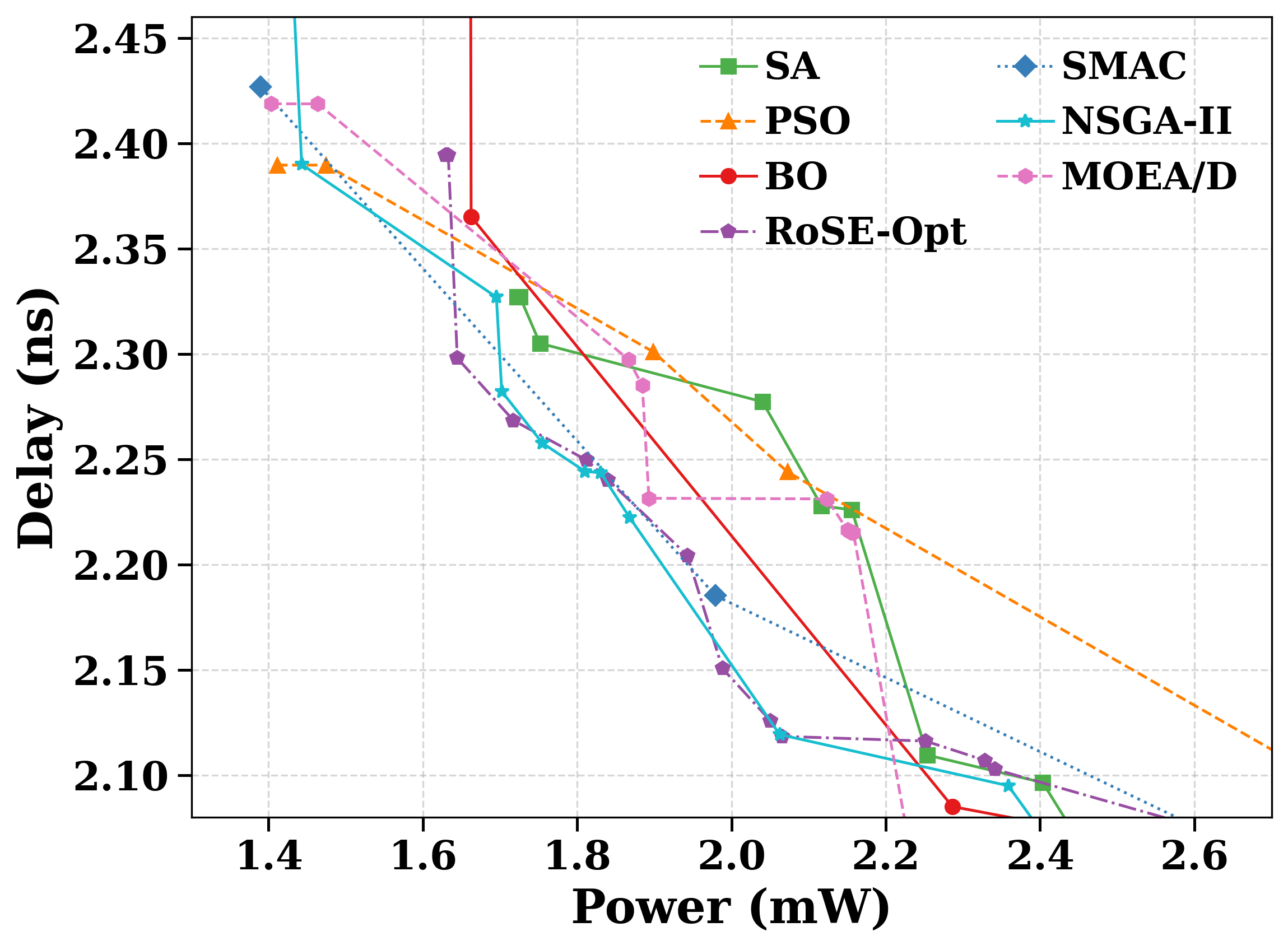} \\ 
        \footnotesize (c) Power-Delay
    \end{minipage}
    \hfill
    \begin{minipage}[b]{0.24\linewidth}
        \centering
        \includegraphics[width=\linewidth]{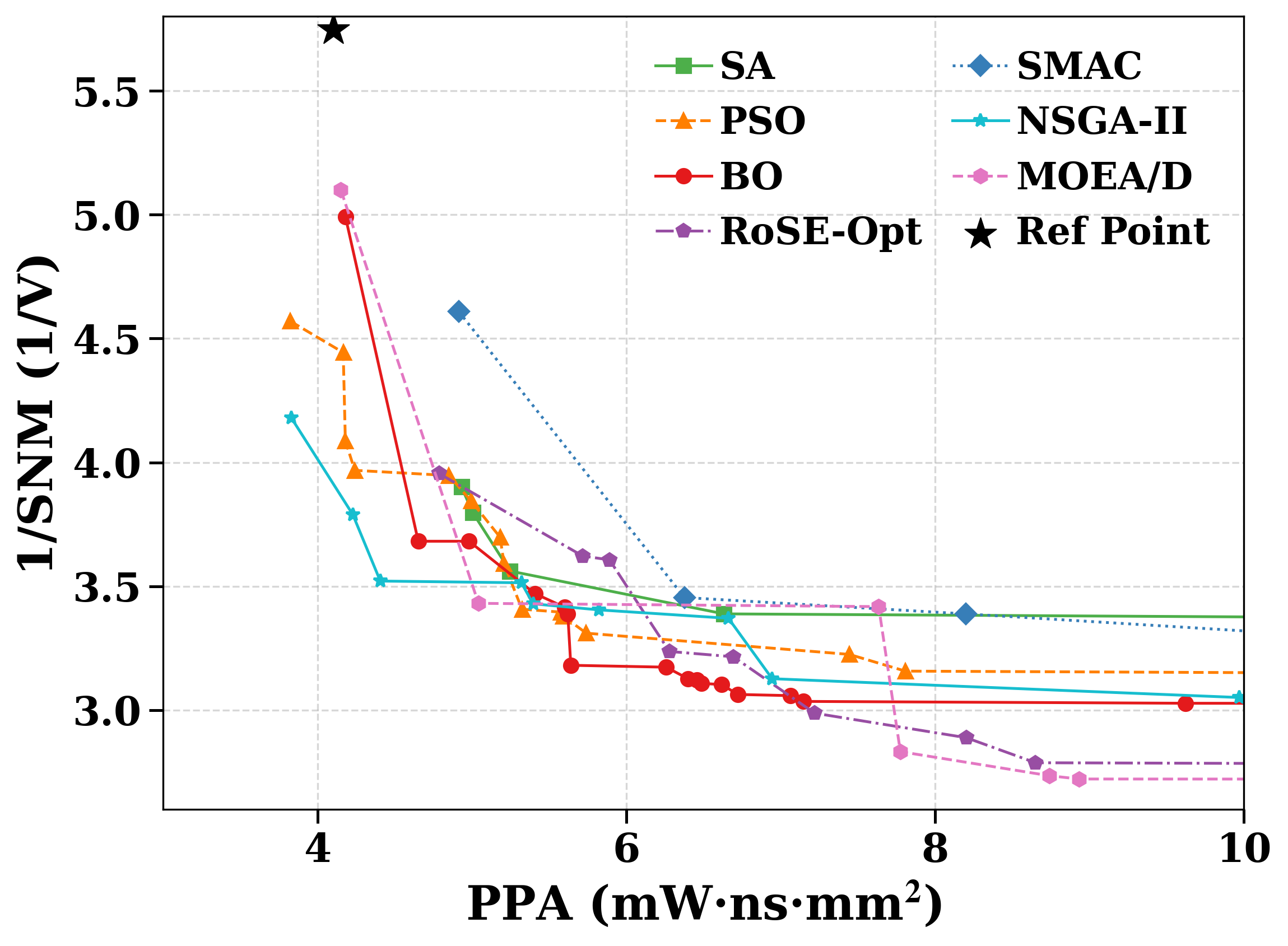} \\ 
        \footnotesize (d) PPA-1/SNM
    \end{minipage}  
    \caption{Overview of optimization algorithm experimental results: (a) FoM convergence curve, (b) Area-SNM comparison, (c) Power-Delay distribution, (d) Trade-off relationship between PPA and stability.}
    \label{fig:opt_results}
\end{figure*}

In this subsection, we evaluate the co-optimization framework using the equivalent circuit model to re-examine the design provided in~\cite{openram} as a starting point. The optimization objective function is defined in Eq.~\eqref{FOM}.

We systematically evaluated seven optimization algorithms. To ensure a fair comparison, all algorithms were allocated an identical budget of 1{,}500 SPICE simulations. The specific algorithmic configurations are as follows:
(i)~\textit{CBO}~\cite{cbo2014}: GP surrogate with constrained EI acquisition;
(ii)~\textit{PSO}~\cite{pso1995}: population size 20, $w{=}0.7$, $c_1{=}c_2{=}1.4$;
(iii)~\textit{SA}~\cite{sa1983}: exponential cooling ($T_0{=}1000$, $T_{\min}{=}10^{-7}$, $\alpha{=}0.98$);
(iv)~\textit{SMAC}~\cite{smac2011}: Random Forest surrogate;
(v)~\textit{RoSE-Opt}~\cite{RoSE_Opt2}: Bayesian optimization with PPO agent;
(vi)~\textit{MOEA/D}~\cite{moead2007}: $h{=}17$, 800 generations, $T{=}20$;
(vii)~\textit{NSGA-II}~\cite{nsga2009}: population size 1{,}200.

Fig.~\ref{fig:opt_results}(a) depicts the FoM convergence trajectories. While NSGA-II and RoSE-Opt exhibit rapid initial convergence, MOEA/D---despite a slower onset---surpasses the FoM threshold of 8.25 after approximately 150 iterations, ultimately attaining the highest solution quality among all evaluated algorithms.

The Pareto fronts are compared in Fig.~\ref{fig:opt_results}(b)--(d). In the Area vs.\ $1/\text{SNM}$ space (Fig.~\ref{fig:opt_results}(b)), MOEA/D consistently achieves the minimum area across the entire stability spectrum, demonstrating superior search efficiency. In the Power vs.\ Delay space (Fig.~\ref{fig:opt_results}(c)), performance differences narrow, though NSGA-II and RoSE-Opt show a slight advantage in the low-power region. For the PPA--Stability trade-off (Fig.~\ref{fig:opt_results}(d)), MOEA/D yields the most robust Pareto front, while NSGA-II leverages its elitist strategy to identify high-performance solutions in the low-PPA regime. All optimized Pareto fronts dominate the reference point, confirming the efficacy of the proposed framework.

Table~\ref{table:opt_results} presents the detailed performance comparison. Under capacity constraints, excessively small arrays incur large area overhead due to the high number of sub-arrays, while excessively large arrays suffer from excessive power and delay penalties. Medium-scale configurations are consistently favored by all algorithms. Compared to the unoptimized baseline, the MOEA/D result achieves the best overall FoM (8.2721): SNM improves by 6.2\% (0.273\,V$\to$0.290\,V), area is reduced by 73.6\%, read and write delays are optimized by 11.3\% and 14.9\% respectively, and peak power is reduced by 42.3\%.

\subsection{Ablation Study: Architecture vs.\ Sizing Contributions}
To disentangle the contributions of architecture selection and transistor sizing, we conduct an ablation study whose results are summarized in Table~\ref{table:ablation}. Starting from the unoptimized baseline (Stage~0), we first apply only the architecture configuration discovered by MOEA/D (Stage~1), and then apply the full co-optimization including sizing (Stage~2).

Architecture optimization (Stage~0$\to$1) dominates the gains.
Consolidating 1024 small $16\times16$ sub-arrays into 128 medium-sized $64\times32$ arrays reduces peripheral overhead---decoder depth, sense amplifiers, precharge drivers, and global routing---by nearly an order of magnitude. The cell-array silicon footprint drops by 68.5\% (from $0.4773\,\text{mm}^2$ to $0.1503\,\text{mm}^2$), which already accounts for 92.9\% of the total area reduction. Read and write delays improve by 9.6\% and 10.7\% respectively, as the shorter bit-line and word-line parasitics of the $64\times32$ tile more than compensate for the logarithmically growing chip-select overhead modeled in Sec.~\ref{sec:penalty}. Peak power falls by 38.8\%, reflecting both the elimination of redundant peripheral activity and the more favorable switching-capacitance profile of the enlarged tile. SNM remains unchanged at 0.273\,V, as expected for a cell-intrinsic metric unaffected by array reorganization. The net effect is a 6.6\% FoM improvement attributable solely to architecture selection.

Transistor sizing (Stage~1$\to$2) provides complementary refinement.
With the architecture fixed, MOEA/D tunes the pull-up/pull-down ratio (narrowing $W_{\mathrm{PD}}$ from $0.205$ to $0.072\,\mu$m), enlarges the pass-gate ($W_{\mathrm{PG}}$: $0.135\to 0.162\,\mu$m), and switches PMOS to high-VTH, delivering an additional 16.6\% area reduction, 5.7\% lower peak power, 6.3\% higher SNM, and 1.9--4.8\% faster access. The larger pass-gate strengthens write-ability without degrading read stability owing to the compensating high-VTH PMOS pull-up. The sizing stage adds a further 1.2\% FoM improvement.

Complementarity of the two stages.
The ablation confirms that neither stage alone can reach the final Pareto-optimal point: architecture optimization establishes a fundamentally more efficient floor-plan baseline (contributing $\sim$84\% of the total FoM gain), while transistor sizing exploits the remaining headroom by shaping the read/write trade-off at the cell level. Crucially, the two stages are not redundant---architecture gains come from reduced peripheral overhead, whereas sizing gains come from cell-level electrical tuning---which is why the joint search outperforms any decoupled approach.

\section{Conclusion}
This paper presents OpenOpt, a co-optimization platform for SRAM architecture and transistor sizing. The equivalent circuit model achieves up to 61.4$\times$ simulation speedup with delay errors within 0.22\% and power errors within 1.68\%. Among seven integrated algorithms, MOEA/D attains the best FoM (8.2721), yielding 73.6\% area reduction, 42.3\% peak power reduction, and 6.2\% SNM improvement. Ablation analysis confirms that architecture selection contributes $\sim$84\% of total FoM gains, while transistor sizing provides complementary refinements, validating the co-optimization strategy.

\vspace{12pt}

\end{document}